# Approximate Kalman Filter Q-Learning for Continuous State-Space MDPs

**Charles Tripp**
Department of Electrical Engineering
Stanford University
Stanford, California, USA
cet@stanford.edu

**Ross Shachter**
Management Science and Engineering Dept
Stanford University
Stanford, California, USA
shachter@stanford.edu

## Abstract

We seek to learn an effective policy for a Markov Decision Process (MDP) with continuous states via Q-Learning. Given a set of basis functions over state action pairs we search for a corresponding set of linear weights that minimizes the mean Bellman residual. Our algorithm uses a Kalman filter model to estimate those weights and we have developed a simpler approximate Kalman filter model that outperforms the current state of the art projected TD-Learning methods on several standard benchmark problems.

## 1 INTRODUCTION

We seek an effective policy via Q-Learning for a Markov Decision Process (MDP) with a continuous state space. Given a set of basis functions that project the MDP state action pairs onto a basis space, a linear function maps them into Q-values using a vector of basis function weights. The challenge in implementing such a projection-based value model is twofold: to find a set of basis functions that allow a good fit to the problem when using a linear model, and given a set of basis functions, to efficiently estimate a set of basis function weights that yields a high-performance policy. The techniques that we present here address the second part of this challenge. For more background on MDPs, see (Bellman, 1957, Bertsekas, 1982, Howard, 1960) and for Q-Learning see (Watkins, 1989).

In **Kalman Filter Q-Learning** (KFQL), we use a Kalman filter (Kalman, 1960) to model the weights on the basis functions. The entire distribution over the value for any state action pair is captured in this model, where more credible assessments will yield distributions with smaller variances. Because we have probability distributions over the weights on the basis functions, and over the observations conditioned on those weights, we can compute Bayesian updates to the weight distributions. Unlike the current state of the art method, Projected TD Learning (PTD), Kalman Filter Q-Learning adjusts the weights on basis functions more when they are less well known and less when they are more well known. The more observations are made for a particular basis function, the more confident the model becomes in its assessment of its weight.

We simplify KFQL to obtain **Approximate Kalman Filter Q-Learning** (AKFQL) by ignoring dependence among basis functions. As a result, each iteration is linear in the number of basis functions rather than quadratic, but it also appears to be significantly more efficient and robust for policy generation than either PTD or KFQL.

In the next section we present KFQL and AKFQL, followed by sections on experimental results, related work, and some conclusions.

## 2 KALMAN FILTER Q-LEARNING

In each time period an MDP has a continuous state $s$ and an action $a$ is chosen from a corresponding set of actions $A_s$. The transition probabilities to a successor state $s'$ are determined by the state action pair $(s, a)$, and the reward $R(s, a, s')$ for that period is determined by the two states and the action. We seek to learn an optimal policy for choosing actions and the corresponding optimal net present value of the future rewards given that we start with state action pair $(s, a)$, the Q-value $Q(s, a)$. To make that analysis tractable it is standard to introduce basis functions $\phi(s, a)$.

Given $n$ basis functions $\phi(s, a)$ for MDP state action pairs $(s, a)$, we learn an $n$-vector of weights $r$ for those basis functions that minimizes the mean Bellman residual. Our prior belief is that $r$ is multivariate normal with $n$-vector mean $\mu$ and $n \times n$ covariance matrix $\Sigma$. We model the Q-value by $Q(s, a) = r^T \phi(s, a)$ with predicted value $\mu^T\phi(s,a)$ and variance $\sigma^2(s,a)$. We then treat the Q-Learning sample update $\nu(s,a)$ as an observation with variance $\epsilon(s,a)$, which we assume is conditionally independent of the prediction given $r$. Under these assumptions we can update our beliefs about $r$ using a Kalman filter. This notation is summarized in Table 1.

**Table 1: Symbol Definitions**

| Symbol | Definition |
| --- | --- |
| $s$ | current MDP state |
| $s'$ | successor MDP state |
| $a \in A_s$ | action available in state s |
| $R(s,a,s')$ | MDP reward |
| $\gamma$ | MDP discount rate |
| $\phi(s,a)$ | basis function values for state action pair $(s,a)$ |
| $r$ | weights on the basis functions that minimize the mean Bellman residual |
| $\mu, \Sigma$ | mean vector and covariance matrix for $r \sim N(\mu,\Sigma)$ |
| $Q(s,a)$ | Q-value for $(s,a)$ |
| $\sigma^2(s,a)$ | variance of $Q(s,a)$ |
| $\nu(s,a)$ | Q-Learning sample update value for $(s,a)$ |
| $\epsilon(s,a)$ | variance of observation $\nu(s,a)$, also called the **sensor noise**. |

Kalman Filter Q-Learning, as discussed here, is implemented using the covariance form of the Kalman filter. In out experiments the covariance form was less prone to numerical stability issues than the precision matrix form. We also believe that dealing with covariances makes the model easier to understand and to assess priors, sensor noise, and other model parameters. While the update equations for the KFQL are standard, the notation is unique to this application. We describe the exact calculations involved in updating the KFQL below.

## 2.1 THE KALMAN OBSERVATION UPDATE

The generic Kalman filter **observation update** equations can be computed for any $r \sim N(\mu,\Sigma)$ with predicted measurement with mean $\mu^T\phi$ and variance $\phi^T\Sigma\phi$. The observed measurement $\nu$ has variance $\epsilon$. First we compute the **Kalman gain** $n$-vector,

$$G = \Sigma\phi(\phi^T\Sigma\phi + \epsilon)^{-1}. \qquad (1)$$

The Kalman gain determines the relative impact the update makes on the elements of $\mu$ and $\Sigma$.

$$\mu \leftarrow \mu + G(\nu - \mu^T\phi) \qquad (2)$$
$$\Sigma \leftarrow (I - G\phi^T)\Sigma \qquad (3)$$

The posterior values for $\mu$ and $\Sigma$ can then be used as prior values in the next update. For in-depth descriptions of the Kalman Filter see these excellent references: (Koller and Friedman, 2009, Russell and Norvig, 2003, Welch and Bishop, 2001).

### 2.1.1 MDP Updates

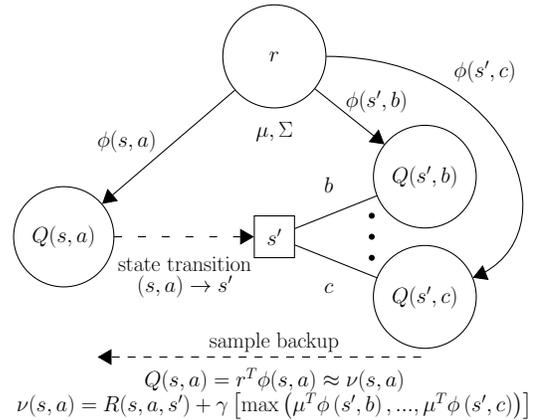

$$Q(s,a) = r^T\phi(s,a) \approx \nu(s,a)$$
$$\nu(s,a) = R(s,a,s') + \gamma\left[\max\left(\mu^T\phi(s',b),...,\mu^T\phi(s',c)\right)\right]$$

**Figure 1:** The Sample Update

To update our distribution for $r$, we estimate the value of a sampled successor state $s'$ and use that value in our observation update. Because this method is similar to the Bellman residual minimizing approximation used in Q-Learning, we will refer to it as the **sample update**. It is shown in Figure 1. To arrive at the update formulation, we start with the Bellman equation:

$$Q(s,a) \approx R(s,a,s') + \gamma \max_{a' \in A_{s'}} Q(s',a') \qquad (4)$$

and then we apply the Q-value approximation $Q(s,a) = r^T\phi(s,a)$, to get:

$$\begin{aligned}
Q(s,a) &= r^T\phi(s,a) & (5) \\
&\approx R(s,a,s') + \gamma E\left[\max_{a' \in A_{s'}} r^T\phi(s',a')\right] & (6) \\
&\approx R(s,a,s') + \gamma \max_{a' \in A_{s'}} \mu^T\phi(s',a') & (7) \\
&= \nu(s,a) & (8)
\end{aligned}$$

There are several approximations implicit in this update in addition to the Bellman equation. First, we assume that the Q-value $Q(s,a)$ can be represented by $r^T\phi(s,a)$. Second, we interchange the expectation and maximization operations to simplify the computation.

Finally, we do not account for any correlation between the measurement prediction and the sample update, both of which are noisy functions of the weights on the basis functions $r$. Instead, we estimate the prediction error $\sigma^2(s,a) = \phi(s,a)^T \Sigma \phi(s,a)$ separately from the sensor noise $\epsilon(s,a)$ as explained below.

### 2.1.2 Computing the Sensor Noise

The sensor noise, $\epsilon(s,a)$, plays a role similar to that of the learning rate in TD Learning. The larger the sensor noise, the smaller the effect of each update on $\mu$. The sensor noise has three sources of uncertainty. First, the sensor suffers from inaccuracies inherent in the model because it is unlikely that the Q-values for all state action pairs can be fit exactly for any value of $r$. Because this source of noise is due to the model used, we will refer to it as **model bias**. Second, there is sensor noise due to the random nature of the state transition because the value of the sample update depends on which new state $s'$ is sampled. We will refer to this source of noise as **sampling noise**. Both model bias and sampling noise are typically modeled as constant throughout the process, and $\epsilon_0$ should be chosen to capture that combination. Finally, there is sensor noise from the assessments used to compute $\nu$ because our assessment of $Q(s',a')$ has variance $\sigma^2(s',a') = \phi(s',a')^T \Sigma \phi(s',a')$. We will refer to this variance as **assessment noise**. Assessment noise will generally decrease as the algorithm makes more observations. All together, the sample update is treated as a noisy observation of $Q(s,a)$ with sensor noise $\epsilon(s,a)$.

We have tried several heuristic methods for setting the sensor noise and found them effective. These include using the variance of the highest valued alternative (Equation 9), using the average of the variances of each alternative (Equation 10), using the largest variance (Equation 11), and using a Boltzmann-weighted average of the variances (Equation 12). Although each of these techniques appears to have similar performance, as is shown in the next section, the average and policy methods performed somewhat better in our experiments.

The policy method:

$$\epsilon(s,a) = \epsilon_0 + \gamma^2 \sigma^2(s', \arg\max_{a' \in A_{s'}} \mu^T \phi(s',a')) \quad (9)$$

The average method:

$$\epsilon(s,a) = \epsilon_0 + \gamma^2 \frac{\sum_{a' \in A_{s'}} \sigma^2(s',a')}{|A_{s'}|} \quad (10)$$

The max method:

$$\epsilon(s,a) = \epsilon_0 + \gamma^2 \max_{a' \in A_{s'}} \sigma^2(s',a') \quad (11)$$

The Boltzmann method:

$$\epsilon(s,a) = \epsilon_0 + \gamma^2 \frac{\sum_{a' \in A_{s'}} \sigma^2(s',a') e^{\frac{\mu^T \phi(s',a')}{\tau}}}{\sum_{a' \in A_{s'}} e^{\frac{\mu^T \phi(s',a')}{\tau}}} \quad (12)$$

Regardless which heuristic method is used, the action selection can be made independently. For example, in our experiments in the next section we use Boltzmann action selection.

Kalman Filter Q-Learning belongs to the family of least squares value models, and therefore each update has complexity $O(n^2)$. This has traditionally given projected TD Learning methods an advantage over least squares methods. Although projected TD Learning is generally less efficient per iteration than least squares methods, each update only has complexity $O(n)$. This means that in practice, multiple updates to the projected TD Learning model can be made in the same time as a single update to the Kalman filter model.

### 2.2 APPROXIMATE KALMAN FILTER Q-LEARNING

We propose **Approximate Kalman Filter Q-Learning** (AKFQL) which updates the value model with only $O(n)$ complexity, the same complexity update as projected TD Learning. The only change in AKFQL is to simply ignore the off-diagonal elements in the covariance matrix. Only the variances on the diagonal of the covariance matrix are computed and stored. This approximation has linear update complexity, rather than quadratic, in the number of basis functions. Although KFQL outperforms AKFQL early on (on a per iteration basis), somewhat surprisingly, AKFQL overtakes it in the long run. We do not fully understand the mechanism of this improvement, but suspect that KFQL is overfitting.

Approximate Kalman Filter Q-Learning's calculations are simplified versions of KFQL's calculations which involve only the diagonal elements of the covariance matrix $\Sigma$. The Kalman gain can be computed by

$$d = \sum_i \phi_i^2 \Sigma_{ii} + \epsilon \quad (13)$$

$$G_i = \frac{\Sigma_{ii} \phi_i}{d}. \quad (14)$$

The Kalman gain determines the relative impact the update makes on the elements of $\mu$ and $\Sigma$.

$$\mu_i \leftarrow \mu_i + G_i(\nu - \mu^T\phi) \quad (15)$$
$$\Sigma_{ii} \leftarrow (1 - G_i\phi_i)\Sigma_{ii} \quad (16)$$

The posterior values for $\mu$ and the diagonal of $\Sigma$ can then be used as prior values in the next update. Finally, to compute the sample noise $\epsilon(s, a)$ we use $\sigma^2(s, a) = \sum_i \phi_i^2(s, a)\Sigma_{ii}$.

### 2.3 THE KFQL/AKFQL ALGORITHM

1 $\mu, \Sigma \leftarrow$ prior distribution over $r$
2 $s \leftarrow$ initial state
3 **while** *policy generation in progress* **do**
4     choose action $a \in A_s$
5     observe transition $s$ to $s'$
6     update $\mu$ and $\Sigma$ using $\nu(s, a)$ and $\epsilon(s, a)$
7     $s \leftarrow s'$
8     **if** *isTerminal(s)* **then**
9         $s \leftarrow$ initial state

**Algorithm 1 :** [Approximate] Kalman Filter Q-Learning

In Kalman Filter Q-Learning (KFQL) and Approximate Kalman Filter Q-Learning (AKFQL) we begin with our prior beliefs about the basis function weights $r$ and we update them on each state transition, as shown in Algorithm 1.

## 3 EXPERIMENTAL RESULTS

Because our primary goal is policy generation rather than on-line performance, our techniques are not meant to balance exploration with exploitation. Instead of testing them on-line, we ran the algorithms off-line, periodically copied the current policy, and tested that policy. This resulted in a plot of the average control performance of the current policy as a function of the number of states visited by the algorithm. In each of our experiments, we ran each test $32 - 50$ times, and plotted the average of the results. At each measurement point of each test run, we averaged over several trials of the current policy.

In our results, we define a state as being **visited** when the MDP entered that state while being controlled by the algorithm. In the algorithms presented within this paper, each state visited also corresponds to one iteration of the algorithm. The purpose of these measurements is to show how the quality of the generated policy varies as the algorithm is given more time to execute. Thus, the efficiency of various algorithms can be compared across a range of running times.

### 3.1 CART-POLE

The **Cart-Pole** process is a well known problem in the MDP and control system literature (Anderson, 1986, Barto et al., 1983, Michie and Chambers, 1968, Schmidhuber, 1990, Si and Wang, 2001). The Cart-Pole process, is an instance of the **inverted pendulum** problem. In the Cart-Pole problem, the controller attempts to keep a long pole balanced atop a cart. At each time step, the controller applies a lateral force, $F$, to the cart. The pole responds according to a system of differential equations accounting for the mass of the cart and pole, the acceleration of gravity, the friction at the joint between the cart and pole, the friction between the cart and the track it moves on, and more.

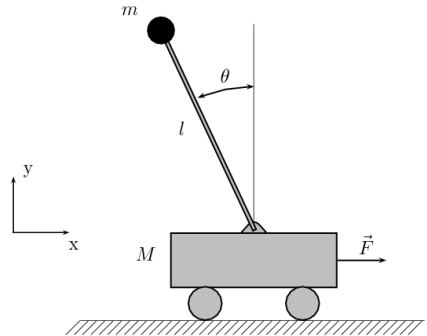

**Figure 2:** The Cart-Pole System

In our experiments, we used the corrected dynamics derived and described by Florian in (Florian, 2007).

In our experiments, we used a control interval of .1 seconds, and the values listed in table 2. At each control interval, the controller chooses a force to apply to the cart in $\{-5N, 0, 5N\}$. Then, this force and an additional force which is uniformly distributed in $[-2N, 2N]$ is also applied to the cart. The controller receives a reward of 1 at each time step except when a terminal state is reached. The problem is undiscounted with $\gamma = 1$. Thus, the number of simulation steps before the pole falls over is equal to the total reward received.

#### 3.1.1 Basis Functions

For the Cart-Pole process, we used a simple bilinear interpolating kernel. For each action, a grid of points is defined in the state space $(\theta, \omega)$. Each point corresponds to a basis function. Thus, every state-action pair corresponds to a point within a grid box with a basis function point at each of the four corners. All basis function values for a particular state are zero except for these four functions. The value of each of the non-zero basis functions is defined by:

$$\tilde{\phi}_i = \left(1 - \left|\frac{\theta - \theta_{\phi_i}}{s_\theta}\right|\right)\left(1 - \left|\frac{\omega - \omega_{\phi_i}}{s_\omega}\right|\right) \quad (17)$$

$$\phi_i = \frac{\tilde{\phi}_i}{\sum_i \tilde{\phi}_i} \quad (18)$$

where $s_\theta = \frac{\pi}{2}$ and $s_\omega = \frac{1}{4}$, and the basis function points are defined for each combination of $\theta_{\phi_i} \in \{-\pi, -\frac{\pi}{2}, 0, \frac{\pi}{2}, \pi\}$ and $\omega_{\phi_i} \in \{-\frac{1}{2}, -\frac{1}{4}, 0, \frac{1}{2}, \frac{1}{4}\}$. This set of basis functions yields $5 \times 5 = 25$ basis functions for each of the three possible actions, for a total of 75 basis functions.

Table 2: Variable definitions for the Cart-Pole process

| VARIABLE | DEFINITION |
| --- | --- |
| $m_c = 8.0 kg$ | Mass of the cart |
| $m_p = 2.0 kg$ | Mass of the pole |
| $l = .5 m$ | Length of the pole |
| $g = 9.81 m/s^2$ | Acceleration of gravity |
| $\mu_c = .001$ | Coefficient of friction between the cart and the track |
| $\mu_p = .002$ | Coefficient of friction between the pole and the cart |
| $\theta$ | Angle of the pole from vertical (radians) |
| $\dot{\theta} = \omega$ | Angular velocity of the pole ($rad/s$) |
| $\ddot{\theta}$ | Angular acceleration of the pole ($rad/s^2$) |
| $F$ | Force applied to the cart by the controller (Newtons) |

Each test on Cart Pole consisted of averaging over 50 separate runs of the generation algorithm, evaluating each run at each test point once. Evaluation runs were terminated after 72000 timesteps, corresponding to two hours of simulated balancing time. Therefore, the best possible performance is 72000.

For projected TD-Learning, a learning rate of $.5\frac{1e6}{1e6+t}$ was used. This learning rate was determined by testing every combination of $s \in \{.001, .005, .01, .05, .1, .5, 1\}$ and $c \in \{1, 10, 100, 1000, 10000, 100000, 1000000, 10000000\}$ for the learning rate, $\alpha_n = s\frac{c}{c+n}$. For KFQL and AKFQL, a sensor variance of $\sigma = .1$, a prior variance of $\Sigma_{ii} = 10000$.

## 3.2 CASHIER'S NIGHTMARE

The **Cashier's Nightmare**, also called **Klimov's problem** is an extension of the multi-armed bandit problem into a queueing problem with a particular structure. In the Cashier's Nightmare, there are $d$ queues and $k$ jobs. At each time step, each queue, $i$, incurs a cost proportional to it's length, $g_i x_i$. Thus, the total immediate reward at each time step is $-g^T x_t$. Further, at each time step the controller chooses a queue to service. When a queue is serviced, its length is reduced by one, and the job then joins a new queue, $j$, with probabilities, $p_{ij}$, based on the queue serviced. Consequently, the state space, $x$, is defined by the lengths of each queue with $x_i$ equaling the length of queue i in state x.

The Cashier's Nightmare has a state space of $\left\langle \begin{array}{c} d \\ k \end{array} \right\rangle = \binom{d+k-1}{k}$ states: one for each possible distribution of the $k$ jobs in the $d$ queues. The instance of the Cashier's Nightmare that we used has $d = 100$ queues, and $k = 200$ jobs, yielding a state space of $\left\langle \begin{array}{c} 100 \\ 200 \end{array} \right\rangle \approx 1.39 \times 10^{81}$ states. This also means that each timestep presents 100 possible actions and each state transition, given an action, presents 100 possible state transitions. In this instance, we set $g_i = \frac{i}{d}$ with $i \in \{1, 2, ..., d\}$ and randomly and uniformly selected probability distributions for $p_{ij}$.

It is worth noting that although we use this problem as a benchmark, the optimal controller can be derived in closed form (Buyukkoc et al., 1983). However, this is a challenging problem with a large state space, and therefore presents a good benchmark for approximate dynamic programming techniques.

### 3.2.1 Basis Functions

The basis functions that we use on Cashier's Nightmare are identical to those used by Choi and Van Roy in (Choi and Van Roy, 2006). One basis function is defined for each queue, and it's value is based on the length of the queue and the state transition probabilities:

$$\phi_i(x, a) = x_i + (1 - \delta_{x_i, 0})(p_{a,i} - \delta_{a,i}) \quad (19)$$

This way, at least the short term effect of each action is reflected in the basis functions.

For projected TD-Learning, a learning rate of $.1\frac{1e3}{1e3+t}$ was used. This learning rate was determined by testing every combination of $s \in \{.001, .01, .1, 1\}$ and $c \in \{1, 10, 100, 1000, 10000\}$ for the learning rate, $\alpha_n = s\frac{c}{c+n}$. For KFQL and AKFQL, a sensor variance of $\sigma = 1$, a prior variance of $\Sigma_{ii} = 20$.

## 3.3 CAR-HILL

The **Car-Hill** control problem, also called the **Mountain-Car** problem is a well-known continuous state space control problem in which the controller attempts to move an underpowered car to the top of a mountain by increasing the car's energy over time. There are several variations of the dynamics of the Car-Hill problem; the variation we use here is the same as that found in (Ernst et al., 2005).

A positive reward was given when the car reached the summit of the hill with a low speed, and a negative reward was given when it ventured too far from the summit, or reached too high of a speed:

$$r(p, v) = \begin{cases} -1 & \text{if } p < -1 \text{ or } |v| > 3 \\ 1 & \text{if } p > 1 \text{ and } |v| \leq 3 \\ 0 & \text{otherwise} \end{cases} \quad (20)$$

The implementation we used approximates these differential equations using Euler's method with a timestep of .001s. The control interval used was .1s, and the controller's possible actions were to apply forces of either $+4N$ or $-4N$. A decay rate of $\gamma = .999$ was used.

Each test on Car-Hill consisted of averaging over 32 separate runs of the generation algorithm, evaluating each run at each test point once. Evaluation runs were terminated after 1000 timesteps, ignoring the small remaining discounted reward.

### 3.3.1 Basis Functions

We used bilinearly interpolated basis functions with an 8x8 grid spread evenly over the $p \in [-1, 1]$ and $v \in [-3, 3]$ state space for each action. This configuration yields $8x8x2 = 128$ basis functions. The prior we used was set to provide higher prior estimates for states with positions closer to the summit:

$$\mu_0(p, v) = \max\left\{0, 1 - \frac{2(1-p)}{3}\right\} \quad (21)$$

For projected TD-Learning, a learning rate of $.1\frac{1e3}{1e3+t}$ was used. This learning rate was determined by testing every combination of $s \in \{.001, .01, .1, 1\}$ and $c \in \{1, 10, 100, 1000, 10000\}$ for the learning rate, $\alpha_n = s\frac{c}{c+n}$. For KFQL and AKFQL, a sensor variance of $\sigma = .5$, a prior variance of $\Sigma_{ii} = .1$.

## 3.4 RESULTS

Kalman Filter Q-Learning has mixed performance relative to a well-tuned projected TD Learning implementation. For the Cart-Pole process (Figure 3), KFQL provides between 100 and 5000 times the efficiency of PTD. However, on the Cashier's Nightmare process (Figure 4) and the Car Hill process (Figure 5), KFQL was plagued by numerical instability, apparently because KFQL underestimates the variance leading to slow policy changes. Increasing the constant sensor noise, $\epsilon_0$, can reduce this problem somewhat, but may not eliminate the issue.

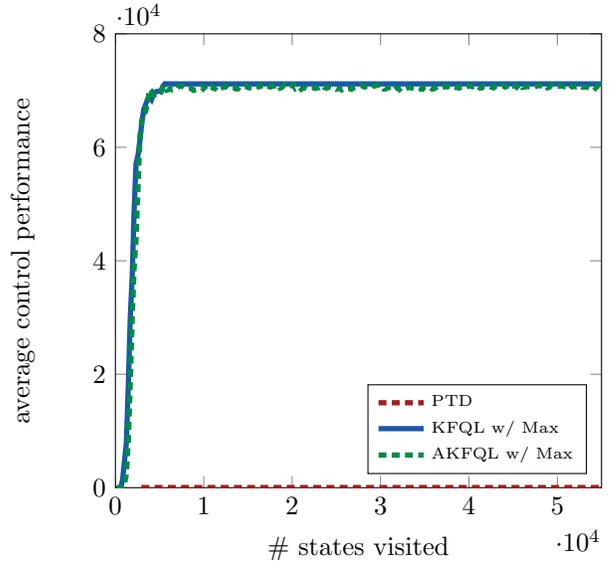

**Figure 3:** PTD, KFQL, and AKFQL on Cart-Pole

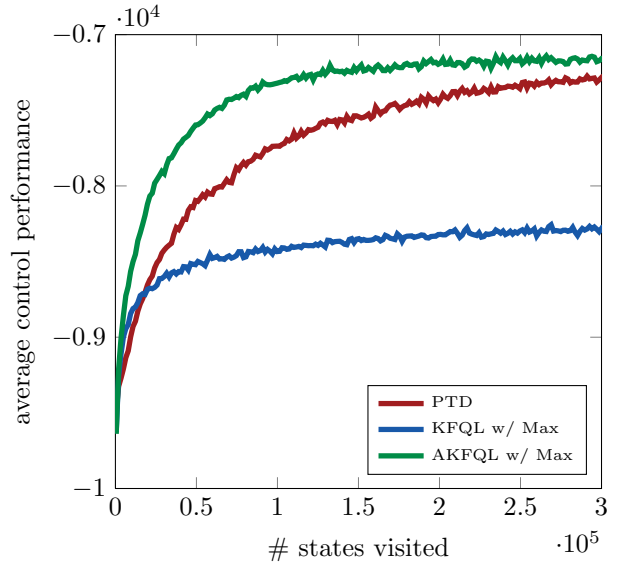

**Figure 4:** PTD, KFQL, and AKFQL on Cashier's Nightmare

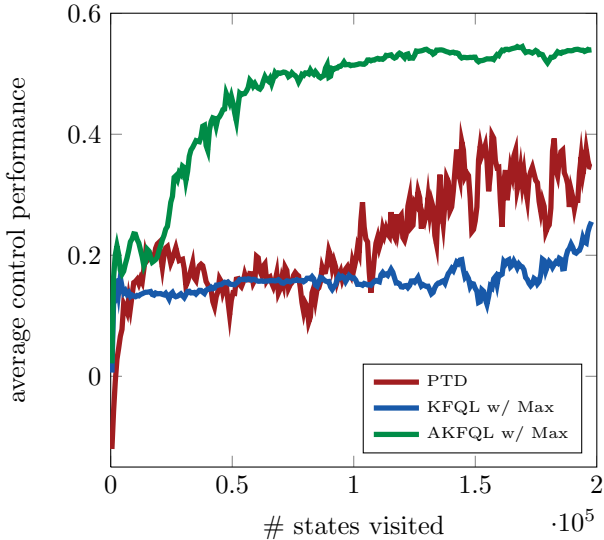

**Figure 5:** PTD, KFQL, and AKFQL on Car-Hill

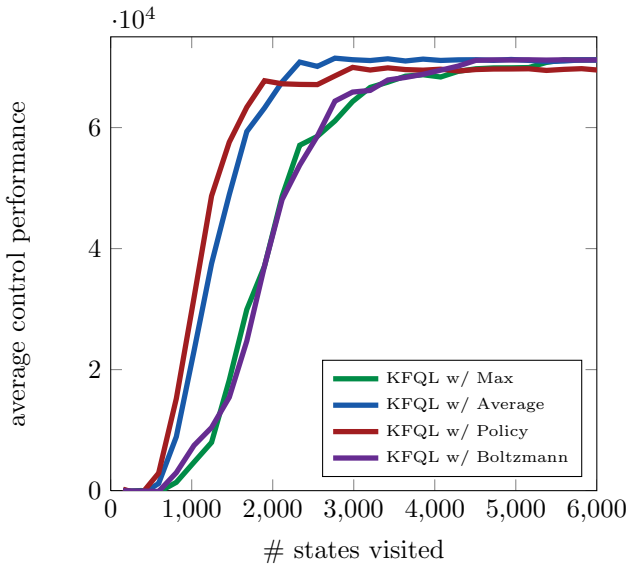

**Figure 6:** Comparison of sensor noise models when using KFQL on Cart-Pole

Approximate Kalman Filter Q-Learning, on the other hand, provides a very large benefit over PTD on all of the tested applications (Figures 3, 4, and 5). Interestingly, in two of the experiments, AKFQL also outperforms KFQL by a large margin, even though the effort per iteration is much less for AKFQL. This is likely due to the fact that AKFQL ignores any dependencies among basis functions. Instead of solving the full least-squares problem (that a Kalman filter solves), AKFQL is performing an update somewhere between the least-squares problem and the gradient-descent update of TD-Learning. This middle-ground provides the benefits of tracking the variances on each basis function's weight, and the freedom from the dependencies and numerical stability issues inherent in KFQL. Additionally, AKFQL's update complexity is linear, just as is PTD's, so it truly outperforms projected TD-Learning in our experiments.

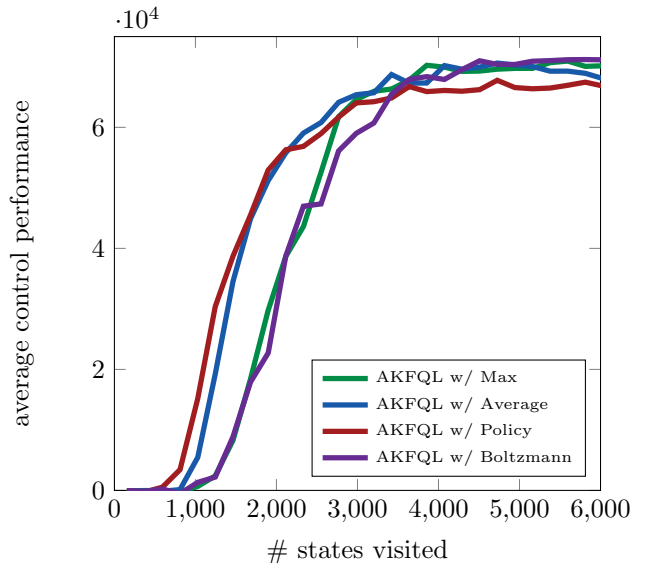

**Figure 7:** Comparison of sensor noise models when using AKFQL on Cart-Pole

The choice of signal noise heuristic has a small effect for the Cart-Pole process under either KFQL (Figure 6) or AKFQL (Figure 7). The average and policy methods appear to work better than the max and Boltzmann methods. The other figures in this section were generated using the max method.

## 4 RELATED WORK

Kalman Filter Q-Learning uses a projected value model, where a set of basis functions $\phi(s, a)$ project the MDP state action space onto the basis space. In these methods, basis function weights $r$ are used to estimate Q-values, $Q(s, a) \approx r^T \phi$, in order to develop high performance policies.

### 4.1 PROJECTED TD LEARNING

**Projected TD Learning** (Roy, 1998, Sutton, 1988, Tsitsiklis and Roy, 1996) is a popular projection-based method for approximate dynamic programming. The version of TD Learning that we consider here is the off-policy Q-learning variant with $\lambda = 0$. Thus, our objective is to find an $r$ such that $Q^*(s,a) \approx r^T \phi(s,a)$ because we are primarily concerned with finding efficient off-line methods for approximate dynamic programming.

The projected TD Learning update equation is:

$$r_{t+1} = r_t + \alpha_t \phi(s_t, a_t) \times \\ \left[ (F\Phi r_t)(s_t, a_t) + w_{t+1} - (\Phi r_t)(s_t, a_t) \right] \quad (22)$$

where $F$ is a weighting matrix and $\Phi$ is a matrix of basis function values for multiple states. We use stochastic samples to approximate $(F\Phi r_t)(x_t)$. The standard Q-Learning variant that we consider here sets:

$$(F\Phi r_t)(s_t, a_t) + w_{t+1} = \\ R(s_t, a_t, s_{t+1}) + \gamma \max_{a \in A_{s_{t+1}}} r^T \phi(s_{t+1}, a) \quad (23)$$

where the state transition $(s_t, a_t) \to (s_{t+1})$ is sampled according to the state transition probabilities, $P(s_t \to s_{t+1} | s_t, a_t)$, of the MDP. Substituting Equation 23 into Equation 22, we arrive at the update equation for this variant of projected TD Learning:

$$r_{t+1} = r_t + \alpha_t \phi(s_t, a_t) \times \\ \left[ \left( R(s_t, a_t, s_{t+1}) + \gamma \max_{a \in A_{s_{t+1}}} r^T \phi(s_{t+1}, a) \right) \\ - (\Phi r_t)(s_t, a_t) \right] \quad (24)$$

The convergence properties of this and other versions of TD learning have been studied in detail (Dayan, 1992, Forster and Warmuth, 2003, Pineda, 1997, Roy, 1998, Schapire and Warmuth, 1996, Sutton, 1988, Tsitsiklis and Roy, 1996). Unlike our Kalman filter methods, PTD has no sense of how well it knows a particular basis functions weight, and adjusts the weights on all basis functions at the global learning rate. As with the tabular value model version of TD learning, the choice of learning rates may greatly impact the performance of the algorithm.

### 4.2 LEAST SQUARES POLICY ITERATION

**Least squares policy iteration** (LSPI) (Lagoudakis et al., 2003) is an approximate policy iteration method which at each iteration samples from the process using the current policy, and then solves the least-squares problem:

$$\text{minimize} \quad ||\hat{A}_t w - \hat{b}_t||_\mu \quad (25)$$

where $\hat{A}$ and $\hat{b}$ are:

$$\begin{aligned}\hat{A} &= \frac{1}{L} \sum_{i=1..L} \phi(s_i, a_i) \left[ \phi(s_i, a_i) - \gamma \phi(s'_i, \pi(s'_i)) \right]^T \\ &\approx \Phi^T \Delta_\mu (\Phi - \gamma P \Pi \Phi)\end{aligned} \quad (26)$$

$$\begin{aligned}\hat{b}_{t+1} &= \hat{b}_t + \phi(s_t, a_t) r_t \\ &\approx \Phi^T \Delta_\mu R\end{aligned} \quad (27)$$

where $P$, $R$, $\Pi$, and $\Phi$ are matrices of probabilities, rewards, policies, and basis function values, respectively.

Solving this least squares problem is equivalent to using KFQL with a fixed-point update method, an infinite prior variance and infinite dynamic noise. LSPI is effective for solving many MDPs, but it can be prone to policy oscillations. This is due to two factors. First, the policy biases $\hat{A}$ and $\hat{b}$, can result in a substantially different policy at the next iteration, which could lead to a ping-ponging effect between two or more policies. Second, information about $\hat{A}$ and $\hat{b}$ is discarded between iterations, providing no dampening to the oscillations and jitter caused by simulation noise and the sampling bias induced by the policy.

### 4.3 THE FIXED POINT KALMAN FILTER

The Kalman filter has also been adapted for use as a value model in approximate dynamic programming by Choi and Van Roy (Choi and Van Roy, 2006). **Fixed point Kalman filter** models the same multivariate normal weights $r$ as KFQL but parametrizes them with means $r_t$ and precision matrix $H_t^{-1}$ such that:

$$r_{t+1} = r_t + \frac{1}{t} H_t \phi(s_t, a_t) \times \\ \left[ (F\Phi r_t)(s_t, a_t) + w_{t+1} - (\Phi r_t)(s_t, a_t) \right] \quad (28)$$

where $(F\Phi r_t)(s_t, a_t) + w_{t+1}$ is the same as in Equation 23, yielding the update rule:

$$r_{t+1} = r_t + \frac{1}{t} H_t \phi(s_t, a_t) \times \\ \left[ \left( R(s_t, a_t, s_{t+1}) + \gamma \max_{a \in A_{s_{t+1}}} r^T \phi(s_{t+1}, a) \right) - \\ (\Phi r_t)(s_t, a_t) \right] \quad (29)$$

where $H_t$ is defined as:

$$H_t = \left[\frac{1}{t}\sum_{i=1}^{t}\phi(s_i,a_i)\phi^T(s_i,a_i)\right]^{-1} \quad (30)$$

Non-singularity of $H_t$ can be dealt with using any known method, although using the psudo-inverse is standard. The choice of an initial $r_0$, $H_0$ and the inversion method for $H$ are similar to choosing the prior means and variance over $r$ in the KFQL. The fixed point Kalman filter represents the covariance of $r$ by a precision matrix rather than the covariance matrix used by KFQL. This difference is significant because KFQL performs no matrix inversion and avoids many non-singularity and numerical stability issues.

### 4.4 APPROXIMATE KALMAN FILTERS

Many methods for approximating and adapting the Kalman filter have been developed for other applications (Chen, 1993). Many of these methods can be used in this application as well. AKFQL is just one simple approximation technique that works well and achieves $O(n)$ instead of $O(n^2)$ complexity updates. Other possible approximation techniques include fixed-rank approximation of $\Sigma$ and particle filtering (Doucet et al., 2001, Gordon et al., 1993, Kitagawa and Gersch, 1996, Rubin, 1987).

## 5 CONCLUSIONS

In this paper we have presented two new methods for policy generation in MDPs with continuous state spaces. The Approximate Kalman Filter Q-Learning algorithm provides significant improvement on several benchmark problems over existing methods, such as projected TD-Learning, as well as our other new method, Kalman Filter Q-Learning. Continuous state MDPs are challenging problems that arise in multiple areas of artificial intelligence, and we believe that AKFQL provides a significant improvement over the state of the art. We believe these same benefits apply to other problems where basis functions are useful, such as MDPs with large discrete sample spaces and mixed continuous and discrete state spaces.

There are a variety of ways this work could be extended. The Kalman filter allows for linear dynamics in the distribution of basis function weights $r$ during transitions from one MDP state action pair to another or from one policy to another. Another possibility is to incorporate more of a fixed point approach, recognizing the dependence between the prediction variance $\sigma^2(s,a)$ and the signal noise $\epsilon(s,a)$ conditioned on $r$. Still another possibility is to use an approximation, ignoring off-diagonal elements like AKFQL, in the traditional fixed point approach. Finally, we could apply a similar Kalman filter model to perform policy iteration directly rather than by Q-Learning.

### Acknowledgements

We thank the anonymous referees for their helpful suggestions.